# A Design Space for Explainable Ranking and Ranking Models

I. Al Hazwani[1,2] , J. Schmid [1] , M. Sachdeva [1] and J. Bernard[1,2]

[1]University of Zurich, Switzerland
[2]Digital Society Initiative, Zürich, Switzerland

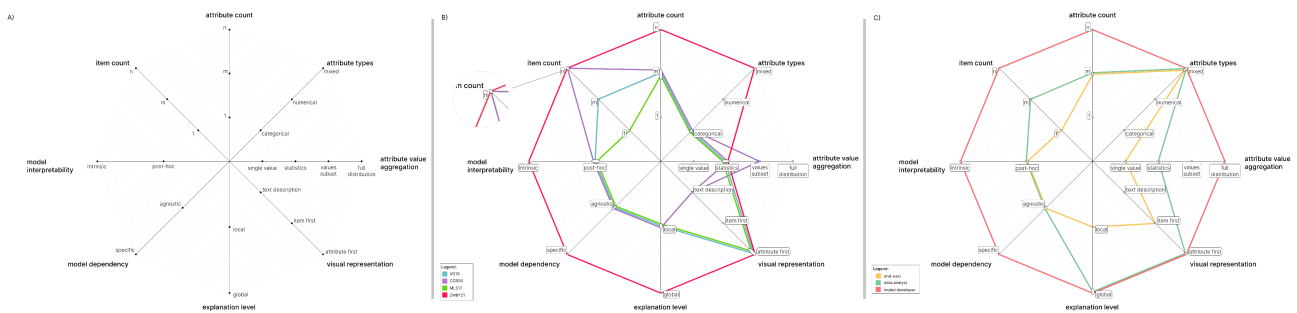

**Figure 1:** *A) Design space for explainable rankings and ranking models composed of 8 dimensions. B) Mapping of four pioneer explainers, according to their explanation characteristics. C) Mapping of three different users groups, exposing different requirements to explainers.*

### Abstract

*Item ranking systems support users in multi-criteria decision-making tasks. Users need to trust rankings and ranking algorithms to reflect user preferences nicely while avoiding systematic errors and biases. However, today only few approaches help end users, model developers, and analysts to explain rankings. We report on the study of explanation approaches from the perspectives of recommender systems, explainable AI, and visualization research and propose the first cross-domain design space for explainers of item rankings. In addition, we leverage the descriptive power of the design space to characterize a) existing explainers and b) three main user groups involved in ranking explanation tasks. The generative power of the design space is a means for future designers and developers to create more target-oriented solutions in this only weakly exploited space.*

## 1. Introduction

Despite the usefulness of item rankings for multi-criteria decision-making tasks, in the real-world end-users often cannot control the ranking creation process by themselves. Existing work reveals that rankings can a) have inherent biases [Cas19], or b) are not human-centered [LV21]. However, only few pioneering works have been presented for explaining item rankings [PP20, ZWB*21]. *Explainable Artificial Intelligence* (XAI) approaches reveal computations behind black-box models [DR20, AB18] such as recommender systems (RS) that generate item rankings. In parallel to explainable RS research, there has been interest from the visualization (VIS) community to assess and analyze item rankings [AL20, LLZ*21, XMT*20] to capture users' feedback [YSJ19, CLK*14], or to support decisions of explanatory chart usage [VHS*17, QRD*20]. In designing explainable item ranking systems, knowledge from different research domains such as XAI, RS, and VIS needs to integrate, with the need for a common language of description. Moreover, different stakeholders with different explainability needs exist, posing the question of how to effectively design explainable ranking systems. In summary, we identify a lack of structure that would a) allow for a joint description of existing solutions and b) guide designers towards novel ranking explanation solutions.

Our primary contribution is a design space of XAI approaches for explaining ranking results and ranking models. The design space is the result of a careful reflection of related work in the three fields XAI, RS and VIS, as well as a synthesis of their inherent concepts and characteristics. To validate the descriptive power [Bea04] of the design space, we map existing explainers into the space, and analyze commonalities and differences across approaches. Our second contribution is the characterization of three main user groups primarily involved in the creation and use of rankings. Using the design space, we discuss main characteristics and design requirements of user groups to ease the abstraction of design targets for future explainers.

## 2. A Design Space for Explainable Rankings

We present a design space for explainable rankings and ranking algorithms, as a result of systematic literature reviews, reflections, as well as a synthesis of the fields *XAI*, *RS*, and *VIS* research.

## 2.1. Abstraction of Three Influential Research Fields

**XAI** : Multiple surveys around explainability have been published providing support from both the model-driven [AS22, MZR21, **?**] and the human-driven perspective [LV21, AS22]. Relevant to our work are approaches for the explanation of rankings, differing in how many items and attributes are considered [VG19, CCR20, MLS17], and as to whether access to the model architecture is supported [ZWB*21]. Finally, XAI differentiates between model-agnostic and model-specific approaches, the latter in the need of incorporating interpretability constraints within the inherent structure and learning mechanisms underlying the black-model [Rai20].

**Recommender Systems**: Interactive RS enable transparency and insight on output rankings [HPV16] using a model-agnostic approach. Approaches differ in their focus on using a) content-based techniques [BVG*15], b) collaborative filtering-based techniques [GOB*10], and c) both [BOH12]. Additionally, approaches differ in visualization techniques used for model interpretability.

**VIS Research**: VIS solutions for the analysis of rankings span from simple forms [KVD*18, ZSHL18] to complex visualization systems [SS04, KR17]. RankVis [PP20] is a pioneer explanation approach that also enables developers to interpret the (learning-to-rank) model, by showing attribute distributions of item subsets. Similar to other VIS approaches, RankVis is attribute-centered, enabling data analysts understand how single [SS04] or multiple [GLG*13, PSTW*17, WDC*18, CL04] focused attributes contribute to the item ranking. In contrast, item-centered attributes show attributes for focused items, e.g., in the context of sensitivity analysis and/or ranking comparisons [GLG*13, XMT*20].

## 2.2. A Design Space for Explainers of Item Rankings

The synthesis of abstractions in *XAI*, *RS*, and *VIS* reveals eight dominating characteristics that form the dimensions of the design space as shown in Figure 1A. Each dimension contains different levels, always with increasing complexity from the center to the outer, useful to a) describe existing explainers and b) generate uncharted room for yet other types of explainers.

**Item Count**: number of explained items, from *1*, to *m*, to *n*.

**Attribute Count**: attributes considered, from *1*, to *m*, to *n*.

**Attribute Type**: attribute types supported by the explainer for explaining the ranking, from *categorical*, *numerical*, to *mixed*.

**Visual Representation**: entails the type of information that is preferred, ranging from *text description*, to *item-first*, to *attribute first*.

**Attribute Aggregation**: characterizes the level of detail used to show distributions of attribute values, from *single value*, to descriptive *statistics*, to *values subset*, to *full distribution*.

**Explanation Level**: explains how a ranking model works, either *local* if it explains decisions for single instances, or *global* if it explains the decision-making mechanism of the entire ranking model.

**Model Dependency**: differentiates whether access to the inherent characteristics of a model architecture is needed to generate an explanation (*model-specific*), or if solely the input and output solves the explanation task (*model-agnostic*).

**Model Interpretability**: an XAI method can be implemented on two different levels: *intrinsic* if the ML model itself provides explanations such as decision trees or logistic regression models, or *post-hoc* in case a helper model is used to mimic the behavior.

## 3. Leveraging the Descriptive Power of the Design Space

### 3.1. Mapping of Existing Explainers

We validate the design space by mapping the four existing XAI approaches for rankings into the space, using the notion of radar charts in combination with metro maps [NMPR07], as Figure 1B shows. It becomes apparent that three approaches have similar characteristics, using a local level of explanation, an agnostic model dependency, and a post-hoc model interpretability [VG19, CCR20, MLS17]. In contrast, the work by Zhuang et al. [ZWB*21] uses intrinsic model interpretability and a global explanation level and is by far the most sophisticated XAI solution observed. What also stands out is that many (combinations of) levels in the design space are weakly populated, given the low number of existing approaches, leaving much room for future implementations. Future developments may be inspired by a more systematic consideration of user groups, as elaborated in the next section.

### 3.2. Characterization and Mapping of User Groups

For the design of explainable item ranking systems, it is essential to take a human-centered approach and align the explanations with user knowledge and user needs [LV21]. Based on existing works and the study of real-world ranking applications, we identify three different user groups with considerable differences, as the design space shows (cf. Figure 1B). With *end users*, *data analysts*, and *model developers*, we characterize user groups by their order of required XAI complexity, from simplistic to most sophisticated.

**End Users** make decisions based on a given ranking, in professional environments and in personal-life situations alike. This user group is in line with the notion of *non-experts* as often used in visualization research [HKPC18], as well as with *AI novices*, as coined in XAI [MZR21]. End users want to understand why an item is located in a specific position and increase their trust in the system, e.g., to make an informed decision in a shopping experience.

**Data Analysts** are interested in the (visual) analysis of multiple items at once, in contrast to end users. Analysts' needs include a model-based explainability using visualization, to draw general conclusions about model characteristics, as well as items and attributes. Data analysts align with *domain experts* in VIS and *data experts* in XAI [MZR21]. Moreover, a related user group that gained attention in VIS is *data journalists* [Her18, KBM21], e.g., those interested in unveiling biases in public ranking models.

**Model Developers** differ from other users by their knowledge on AI and RS, and by their focus on model-building. Model developers, called *AI experts* in XAI [MZR21], have an interest in the interpretability of rankings, ranking models, and their behavioral characteristics in particular, e.g., for debugging and refinement purposes. They want detailed explanations (intrinsic model interpretability and specific model dependency) and often rely on RS techniques to create the underlying ranking system [HPV16].

## 4. Conclusion and Future Work

We proposed a design space for explainable ranking by merging knowledge from XAI, RS, and VIS. We validated our design space by mapping previous work on explainable rankings and user needs. Future work includes a) expanding the design space towards user interaction, b) leveraging the design space to design human-centered XAI ranking systems, and c) cross-cutting the design space with industry human-AI guidelines [WWP*20].